\documentclass[...]{article}

\usepackage{arxiv}

\usepackage[utf8]{inputenc} 
\usepackage[T1]{fontenc}    
\usepackage{hyperref}       
\usepackage{url}            
\usepackage{booktabs}       
\usepackage{amsfonts}       
\usepackage{nicefrac}       
\usepackage{microtype}      
\usepackage{lipsum}
\usepackage{graphicx}
\graphicspath{ {./images/} }
\usepackage{graphicx}%
\usepackage{multirow}%
\usepackage{amsmath,amssymb,amsfonts}%
\usepackage{amsthm}%
\usepackage{mathrsfs}%
\usepackage[title]{appendix}%
\usepackage{textcomp}%
\usepackage{manyfoot}%
\usepackage{booktabs}%
\usepackage{algorithm}%
\usepackage{algorithmicx}%
\usepackage{algpseudocode}%
\usepackage{listings}%

\usepackage[table]{xcolor}

\title{Modified Adaptive Tree-Structured Parzen Estimator for Hyperparameter Optimization}

\author{
 Szymon Sieradzki \\
  Faculty of Mathematics and Information Science\\
  Warsaw University of Technology\\
  Koszykowa 75, Warsaw, Poland \\ \\
  \texttt{szymon.sieradzki2.stud@pw.edu.pl} \\
   \And
 Jacek Ma{\'n}dziuk \\
  Faculty of Mathematics and Information Science\\
  Warsaw University of Technology\\
  Koszykowa 75, Warsaw, Poland \\ \\
  Faculty of Computer Science \\
  AGH University of Krakow \\
  A. Mickiewicza 30, Krakow, Poland \\ \\
  \texttt{jacek.mandziuk@pw.edu.pl} \\
}

\begin{document}
\maketitle
\begin{abstract}
In this paper we review hyperparameter optimization methods for machine learning models, with a particular focus on the Adaptive Tree-Structured Parzen Estimator (ATPE) algorithm. 
We propose several modifications to ATPE and assess their efficacy on a diverse set of standard benchmark functions.
%
%
Experimental results demonstrate that the proposed modifications 
significantly improve the effectiveness of ATPE hyperparameter optimization on selected benchmarks, 
a finding that holds practical relevance for their application 
in real-world machine learning / optimization tasks.
\end{abstract}


\section{Hyperparameter Optimization (HPO)}

In machine learning, the performance of a model heavily depends on the correct choice of hyperparameters, such as the learning rate, the number of layers in a neural network, or specific regularization techniques. These hyperparameters form a multidimensional space where some dimensions are continuous (e.g., the learning rate), while others are discrete (e.g., the number of network layers).

The task of Hyperparameter Optimization (HPO) aims to find the best combination of these hyperparameters by searching this space in a way that optimizes a predefined objective function. In supervised learning, this function is usually a loss function, which quantifies an error between the predictions of the model and the true values. HPO is applicable across a wide range of machine learning models, as most optimization techniques are agnostic to the underlying model type. The core requirement for any HPO algorithm is to define the hyperparameter space and the objective function. However, HPO presents specific challenges that separate it from other optimization problems, making it a unique area in the field.

\subsection{Key Characteristics of HPO}

A key aspect that distinguishes HPO from most other optimization problems is high computational cost of evaluating the objective function. Each evaluation of the objective function requires training the considered machine learning model from scratch, which is often the most time-consuming part of the optimization process. As a result, when designing HPO algorithms, the focus is less on the internal computational efficiency of the optimizer but rather on minimizing the number of objective function evaluations, while still maintaining good predictive performance.

The structure of the hyperparameter space itself introduces further challenges, as it often spans dozens of dimensions, making the search inherently complex. In addition, certain hyperparameters may depend on others, leading to complex interactions. For instance, in a neural network, if fewer than $n$ layers are used, the number of neurons in the $n$-th (or further) layers become irrelevant to the objective function. More advanced HPO algorithms that can account for such dependencies tend to perform better with fewer evaluations by focusing on the most impactful dimensions of the hyperparameter space.

\subsection{Historical Approaches to HPO}

Over the past decade, several approaches to HPO that gained popularity had been introduced. These methods range from simple model-free search strategies to more sophisticated techniques such as evolutionary algorithms or probabilistic models. Each of these methods has unique characteristics, making it difficult, if at all possible, to define a single best approach to HPO. Instead, each technique tends to excel in specific scenarios, depending on the class of machine learning models or the nature of the task.

Model-free methods, such as grid search or random search, are widely used due to their simplicity, but they can be inefficient, especially in high-dimensional spaces. Evolutionary algorithms, inspired by natural selection, have been explored as a more dynamic alternative, offering better performance by simulating evolution to refine hyperparameters. Probabilistic models, such as Bayesian optimization, have also been successfully applied to HPO, allowing the search to be guided by uncertainty and past observations, which makes them more sample-efficient.

All of these methods possess unique strengths, making them more suitable for certain types of machine learning models and tasks. Thus, selecting the optimal HPO method is highly problem-dependent, with no universally superior solution.

The simplest approach to HPO is model-free hyperparameter search, which is independent of the objective function's values from previous iterations. Two popular methods in this category are \textbf{random search}, which selects hyperparameters randomly according to a predefined distribution, and \textbf{grid search}, which systematically evaluates a fixed grid of hyperparameter combinations.

\textbf{Model-free methods} are widely used due to their simplicity and effectiveness in scenarios where the model can be trained quickly. They focus on exploration and avoiding local optima, but converge slower to a global optimum. While grid search might seem more structured, it was demonstrated that random search is often superior in high-dimensional spaces, as grid search tends to waste evaluations on less relevant hyperparameters~\cite{bergstrarandom}.

\textbf{Evolutionary algorithms} are among the most popular methods for heuristic-free optimization and have been widely adopted in HPO. Like model-free search, they are efficient in terms of optimizer runtime. However, they focus more on exploitation, allowing faster convergence to local optima. These methods are particularly useful when the cost of training a model is low, and the available budget permits multiple evaluations of the candidate hyperparameter selections~\cite{loshchilov2016cmaes}. 

\textbf{Probabilistic models}, though not the most actively developed, remain one of the most effective approaches to HPO. They estimate probabilities of achieving certain objective function values in different regions of the hyperparameter space. Their key advantage lies in delivering strong results with relatively few evaluations (up to several thousand), making them an ideal approach when training a model is expensive. The most popular probabilistic methods in HPO are Gaussian Process Estimation (GPE)~\cite{snoek2012practicalbayesianoptimizationmachine} and Tree-Structured Parzen Estimator (TPE)~\cite{10.5555/2986459.2986743}, presented in detail in Section~\ref{sec:TPE}. An enhanced version of the TPE model, the Adaptive Tree-Structured Parzen Estimator (ATPE)~\cite{arsenault2023learning}, which is the focus of our proposed modifications, is presented in Section~\ref{atpe-sect}.

\section{Related work}

This section provides an overview of prominent methods applied to hyperparameter optimization (HPO), specifically focusing on evolutionary algorithms and probabilistic models. Evolutionary algorithms are widely used in HPO due to their ability to explore large, complex search spaces without requiring gradient information. These algorithms simulate natural evolutionary processes to iteratively improve hyperparameter configurations, making them particularly effective in problems where objective function evaluations are computationally manageable.

Probabilistic models, on the other hand, offer a more refined approach by predicting objective function values based on prior observations. GPE directly models the objective function with a Gaussian process, providing both predictions and uncertainty estimates, which allows it to balance exploration and exploitation effectively. TPE, in contrast, models the probability of high and low objective values separately, offering computational efficiency and flexibility in high-dimensional search spaces.

By comparing these approaches, this section highlights the unique advantages and challenges associated with evolutionary and probabilistic methods, setting the stage for an in-depth analysis of the Adaptive Tree-Structured Parzen Estimator (ATPE) algorithm in subsequent sections.

\subsection{Evolutionary approach}

One of the first successful applications of evolutionary algorithms in HPO utilized the CMA-ES algorithm \cite{loshchilov2016cmaes}, originally introduced in \cite{6790628}. Although initially seen as effective only for local optimization, it was later shown to yield satisfactory results in global optimization tasks as well \cite{10.1007/978-3-540-30217-9_36}.

In their study, Loshchilov and Hutter \cite{loshchilov2016cmaes} compared CMA-ES with leading probabilistic HPO algorithms of Adam and AdaDelta models in the MNIST digit recognition task~\cite{mnist}. Their findings align with broader observations throughout our work, indicating that while probabilistic methods, such as TPE, are more effective with a limited number of function evaluations, evolutionary methods like CMA-ES often outperform them when a larger evaluation budget is available. This trend highlights the trade-offs inherent in choosing an HPO strategy based on the computational resources and specific requirements of the task.

Yeh et al.~\cite{YEH2023109076} applied Simplified Swarm Optimization (SSO) to optimize hyperparameters of the LeNet convolutional neural network architecture.
SSO, a variation of swarm algorithms introduced in 2009~\cite{YEH20099192}, iteratively adjusts individuals in a population based on their proximity to the best-performing individuals. In this study, the authors introduced modifications to SSO tailored specifically to the hyperparameter space of LeNet, such as ensuring a non-increasing sequence of neuron counts across feature-extracting layers.

They experimentally demonstrated that across several tasks the optimized hyperparameters achieved better results than those originally proposed by LeCun.
The results on three classification tasks — MNIST~\cite{mnist}, Fashion-MNIST~\cite{fashionmnist}, and CIFAR10~\cite{cifar10} — show that the SSO-optimized version of LeNet (SSOLeNet) achieved superior accuracy and faster inference times compared to the original LeNet architecture. At the same time, training times were generally higher for SSOLeNet, which should be attributed to the repeated training required for hyperparameter optimization, reflecting the computational cost of the entire optimization process.

Japa et al.~\cite{10128116} proposed the use of the Biased Random Key Genetic Algorithm (BRKGA) for hyperparameter optimization and demonstrated that BRKGA could outperform both CMA-ES and probabilistic methods, given a sufficiently long optimization timeframe.

RKGA, originally proposed in~\cite{Bean1994}, operates as a genetic algorithm that uses random keys within a continuous range, enhancing flexibility across different problem types. BRKGA~\cite{GoncalvesResende2011} extends RKGA by increasing the selection bias toward top-performing individuals, which potentially explains its comparative advantage by strengthening exploration.

Japa et al. further adapted BRKGA into HyperBRKGA, incorporating local methods for exploiting neighboring solutions at the end of each iteration. Specifically, they introduced two approaches: random walk and Bayesian walk. In the Bayesian variant, a Gaussian Process Estimator (GPE) model is constructed to predict the best nearby solutions, enhancing local exploitation while potentially sacrificing the ability to escape from local optima. The authors also implemented a data reduction mechanism for the GPE training set, which discards similar and suboptimal individuals to maintain manageable computation times.

Throughout various classification tasks, HyperBRKGA achieved significantly better performance metrics than both evolutionary and Bayesian optimization algorithms. However, due to the lack of information on the number of function evaluations, it remains unclear whether HyperBRKGA would outperform Bayesian optimization methods in scenarios with limited function calls.

The authors' concept of integrating evolutionary and probabilistic algorithms may lay the foundation for a new class of HPO methods. Future enhancements could include using TPE in place of GPE and a more advanced data reduction algorithm, such as correlation-based methods, similarly to the ones implemented in ATPE.

A different approach to HPO are shyperheuristic methods that rely on self-adaptation and hybridization of several algorithms with the focus on the optimal selection of the constituent methods along with their parameterization. The related literature is broad, in particular regarding self-adaptation and hybridization of the PSO method~\cite{Okulewiczetal2022,okulewiczetal2020}.  

\subsection{Probabilistic model based approach}
\label{sec:TPE}

Probabilistic models have become a cornerstone of hyperparameter optimization, offering a distinct advantage over simpler search methods through their ability to model uncertainty in the objective function. Unlike model-free approaches that sample hyperparameters without considering past outcomes, probabilistic models estimate the likelihood of obtaining promising results in unexplored regions of the hyperparameter space. This targeted approach makes them particularly effective when evaluations are computationally expensive.

A common framework for implementing probabilistic models in HPO is Sequential Model-Based Optimization (SMBO)~\cite{10.1007/978-3-642-25566-3_40}. In SMBO, a probabilistic model is iteratively refined with each evaluation, balancing exploration of new hyperparameter configurations with exploitation of known high-performing areas. This dynamic adaptation allows SMBO methods, such as GPE and TPE, to achieve optimal results with fewer evaluations compared to traditional techniques.

\textbf{Gaussian Process Estimator (GPE)~\cite{snoek2012practicalbayesianoptimizationmachine}} is widely used in hyperparameter optimization due to its ability to model the uncertainty of the objective function and to make accurate predictions with a limited number of evaluations. GPE assumes a joint Gaussian distribution for any set of observed function values, allowing it to predict objective values at unobserved points based on prior data, thereby guiding and maintaining the search process more effectively.

GPE’s main advantage lies in its capacity to balance exploration and exploitation. By calculating both the predicted function values and their associated uncertainty, GPE can identify regions with high optimization potential while sampling points with high uncertainty to discover new information. This approach reduces the need for excessive evaluations, making GPE particularly beneficial in scenarios where each function evaluation is computationally intensive.

However, a limitation of GPE is its computational complexity, which scales cubically with the number of data points ($\mathcal{O}(n^3)$). This is due to the need to invert the covariance matrix, which becomes more costly as the dataset of past evaluations grows. Consequently, for large datasets, GPE often requires data reduction techniques to remain efficient, such as pruning less informative points or clustering data to reduce computational load.

GPE has been effectively utilized in HPO for various machine learning models. Snoek et al.~\cite{snoek2012practicalbayesianoptimizationmachine} demonstrated the superiority of Bayesian optimization with GPE over traditional model-free methods like random and grid search in tuning deep neural networks and other complex models. Similarly, Hutter et al.~\cite{10.1007/978-3-642-25566-3_40} applied GPE-based methods to optimize hyperparameters in Support Vector Machines (SVMs) and random forests, showing that, compared to other approaches, GPE could achieve better performance with fewer evaluations. These studies underscore GPE’s effectiveness, especially in high-dimensional, continuous search spaces with constrained evaluation budgets, as observed throughout this paper.

GPE is not only used in traditional machine learning models but has also demonstrated its scalability and effectiveness in optimizing hyperparameters for complex high-dimensional tasks. For instance, Wang et al.~\cite{wang2013bayesian} proposed an approach using random embeddings to extend the applicability of Gaussian Processes to high-dimensional optimization problems. This technique reduces the effective dimensionality of the search space, enabling Gaussian Processes to scale to tasks with hundreds of dimensions while retaining their predictive accuracy and efficiency.

Despite its strengths, the computational complexity of GPE remains a challenge, especially as datasets grow larger. Techniques such as sparse Gaussian Processes~\cite{titsias09a} and random embeddings~\cite{wang2013bayesian} have been introduced to address this limitation, making GPE a viable choice for large-scale hyperparameter optimization. These advancements ensure that GPE remains relevant for modern applications requiring high-dimensional search spaces and limited evaluation budgets.

\textbf{Tree-Structured Parzen Estimator (TPE)~\cite{10.5555/2986459.2986743}} is a popular probabilistic approach in HPO, designed to efficiently model the objective function by constructing separate density estimators for "good" and "bad" hyperparameter values. Unlike GPE, which models the objective function directly, TPE models it indirectly by estimating two conditional probability distributions: one for configurations that yield high objective values and another one for those leading to lower values. This distinction enables TPE to focus on promising regions of the hyperparameter space.

TPE’s advantage lies in its flexibility and computational efficiency, especially for high-dimensional or complex hyperparameter spaces where GPE might struggle. TPE has a linear time complexity $\mathcal{O}(n)$ with respect to the number of evaluations, allowing it to scale more efficiently than GPE, particularly as the dataset grows. Additionally, TPE can handle both continuous and categorical variables, making it versatile for a variety of HPO tasks.

This approach, as shown in~\cite{10.5555/2986459.2986743}, often outperforms grid and random search, especially in high-dimensional spaces where only a subset of hyperparameters significantly influences performance. Its ability to prioritize exploration of promising regions makes it particularly useful in tasks where a large search space and a limited number of evaluations are an issue. In practice, TPE has been successfully applied in tuning hyperparameters for models such as deep neural networks and SVMs, demonstrating strong performance across diverse optimization problems~\cite{bergstra2012makingsciencemodelsearch}.

Recent work has highlighted TPE's adaptability to modern optimization problems. For instance, it has been successfully applied in neural architecture search (NAS)\cite{zoph2017neuralarchitecturesearchreinforcement}, where the search space complexity significantly exceeds traditional hyperparameter tuning tasks. Moreover, TPE has been integrated into AutoML pipelines, such as Auto-sklearn\cite{NIPS2015_11d0e628}, demonstrating its compatibility with broader machine learning automation workflows. The method's ability to model separate density distributions has also been extended to reinforcement learning tasks, offering an efficient means of hyperparameter tuning in dynamic environments~\cite{Henderson_Islam_Bachman_Pineau_Precup_Meger_2018}.

\section{Adaptive Tree Structured Parzen Estimator (ATPE)}
\label{atpe-sect}

The Adaptive Tree-Structured Parzen Estimator (ATPE)~\cite{arsenault2023optimizing} was introduced by Bradley Arsenault who subsequently refined the algorithm in~\cite{arsenault2023learning}. This chapter provides an in-depth look at the key components of the latter version of ATPE (heneceforth referred to as ATPE), since this algorithm serves as the foundation for the research presented in the following chapters.

At its core, ATPE relies on an unmodified TPE, with additional components that affect the algorithm’s input. Specifically, Arsenault's implementation, which builds on the TPE functionality available in the Hyperopt library~\cite{bergstra2013making}, includes mechanisms for:
\begin{itemize}
    \item reducing the number of samples,
    \item restricting certain hyperparameter values,
    \item optimizing both TPE and ATPE hyperparameters.
\end{itemize}

\subsection{Sample Reduction}
\label{7-atpe-filt}

In classical SMBO, all historical samples are used to build the model at each iteration. ATPE modifies this process by introducing a filtering mechanism for the samples fed into the model. This concept was inspired by a TPE modification proposed in~\cite{bergstra2012makingsciencemodelsearch}, where the authors introduced a weighting of historical samples based on their quality, allowing the upper and lower sample sets in TPE to better reflect the distribution of “effective” and “ineffective” samples (for example, by reducing the weights assigned to outliers).

ATPE builds on the observation that weighting samples is a generalization of their elimination (elimination is equivalent to assigning a weight of zero to a sample), but the elimination process is easier to manage. Instead of predicting a numerical value for a sample, one only needs to classify it. In summary, ATPE the following filtering schemes:

\begin{itemize}
    \item no filtering,
    \item random filtering – samples are randomly eliminated, with their percentage being a model parameter,
    \item age-based filtering – the probability of eliminating a sample depends on how long it has been in the history and on a model-defined multiplier,
    \item objective-based filtering – the probability of eliminating a sample is influenced by its objective function value and a model-defined multiplier.
\end{itemize}

The choice of a filtering method is the one of ATPE parameters.

\subsection{Hyperparameter Blocking}

ATPE introduces the concept of hyperparameter blocking, which optimizes the objective function within a reduced hyperparameter space. In this approach, some hyperparameters are fixed to constant values, simplifying the TPE search space during optimization.

This blocking process requires two main components:
\begin{itemize}
    \item selection of hyperparameters to be blocked,
    \item selection of the values for these blocked hyperparameters.
\end{itemize}

The selection of hyperparameters to be blocked in ATPE consists of two phases:
\begin{enumerate}
    \item correlation-based selection (cf. Section \ref{7-atpe-corr-sect}),
    \item from the hyperparameters chosen in the first phase, a subset is selected randomly based on one of the two criteria:
    \begin{itemize}
        \item a fixed probability, set by a model parameter,
        \item a probability based on weighted correlation (see Section \ref{7-atpe-corr-sect}), scaled by a multiplier which is a model parameter.
    \end{itemize}
\end{enumerate}

For assigning values to blocked hyperparameters, ATPE provides two methods:
\begin{itemize}
    \item random selection – the value is randomly chosen from all samples in the history,
    \item elite selection – the value is randomly chosen from a percentage of the top-performing samples based on objective function values, with this elite percentage being a model parameter.
\end{itemize}

\subsubsection{Correlation-Based Selection}
\label{7-atpe-corr-sect}

In the correlation-based selection, only non-categorical hyperparameters are considered, as correlation can meaningfully capture relationships with the objective function only in continuous domains.

For each non-categorical hyperparameter, Spearman’s rank correlation with the objective function is calculated:
\begin{equation}
\rho(x) = 1 - \frac{6 \sum_{i=1}^n d_i(x)^2}{n(n^2-1)}
\label{eq:Spearman}
\end{equation}
where $n$ is the number of samples, and $d_i$ is the rank difference for the $i$-th sample after sorting. Using Spearman's correlation is essential here, as it is less affected by gradient differences between hyperparameter values and objective function values than Pearson's correlation.

The weighted correlation is then calculated as:
\begin{equation}
\rho'(x) = |\rho(x)|^\alpha
\end{equation}
where $\alpha$ is a model parameter.

The hyperparameters are sorted in descending order based on $\rho'(x)$, forming the sequence $a_{i=1,..,K}$.

Next, the weighted correlation sum of all hyperparameters under consideration is computed and scaled by the absolute value of the threshold $\beta$, a model parameter:
\begin{equation}
T = \left(\sum_{i=1}^K \rho'(a_i)\right) \cdot |\beta|
\end{equation}

If $\beta \geq 0$, the hyperparameters with the highest weighted correlation are selected for blocking until their cumulative correlation sum reaches the threshold:
\[
W = \{i: G(i) \leq  T, \quad 1 \leq i \leq K\}
\]
where
\[
G(i) = \rho'(a_{i}) + G(i-1), \quad G(0) = 0 \land 0 \leq i \leq K
\]

If $\beta < 0$, hyperparameters with the lowest weighted correlation are selected until their cumulative correlation sum reaches the threshold:
\begin{equation}
W = \{i: L(i) \leq T,  \quad 1 \leq i \leq K\}
\end{equation}
where
\begin{equation}
L(i) = \rho'(a_{i}) + L(i+1), \quad L(K+1) = 0 \land 1 \leq i \leq K + 1
\end{equation}

\subsection{Optimization of ATPE Parameters}
\label{7-atpe-optim}

The ATPE performance, as presented above, is highly dependent on the choice of its parameters. To streamline the proces of selection of their values, ATPE implements a machine learning (ML) model. Similartly to TPE, ATPE is an SMBO algorithm, meaning that each iteration only shares the sample history with the next one. This allows ATPE parameters to be adjusted dynamically between iterations.

\begin{figure}
\centering
\includegraphics[width=0.65\textwidth]{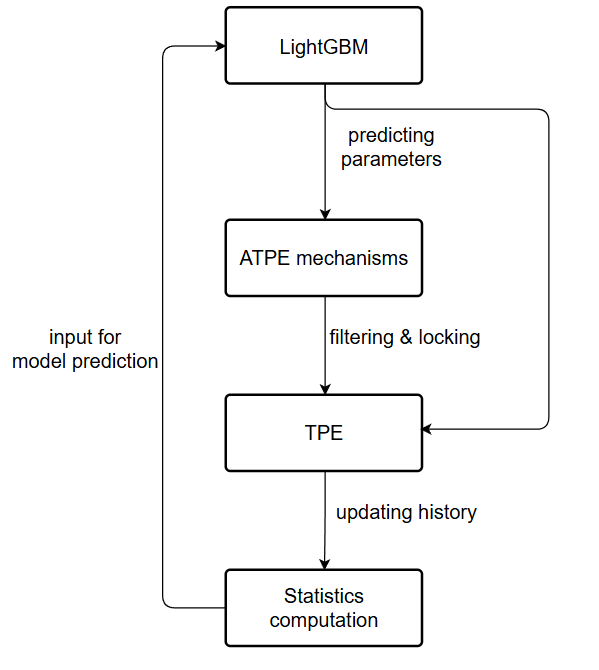}
\caption{Information flow in ATPE}
\label{fig:atpe_flow}
\end{figure}

In each iteration, a machine learning model determines ATPE's parameter values based on the statistics of the previous iteration. An optimization step is then performed, new statistics are collected, and the process repeats (see Figure~\ref{fig:atpe_flow}). These statistics are discussed in detail in Section~\ref{7-atpe-stats}.

The ML model used for parameter prediction is the widely recognized LightGBM, known for its efficiency and scalability in gradient boosting \cite{ke2017lightgbm}. The prediction process is cascaded: the first parameter is predicted based solely on statistics, the second one, based on statistics and the value of the first parameter, and so on. Each parameter has its own LightGBM model, with the training process described in Section~\ref{7-atpe-teaching}.

\subsection{ATPE Statistics}
\label{7-atpe-stats}

ATPE computes a consistent set of statistics for both correlation and objective function values:
\begin{itemize}
    \item ratio of the highest value to the 25th percentile value,
    \item ratio of the highest value to the 50th percentile value (median),
    \item ratio of the highest value to the 75th percentile value,
    \item kurtosis of the values,
    \item ratio of the 25th percentile value to the 5th percentile value,
    \item skewness of the values,
    \item ratio of the standard deviation to the highest value.
\end{itemize}

These statistics are calculated across various sample ranges:
\begin{itemize}
    \item all samples,
    \item the last 10 samples,
    \item the last 15 samples,
    \item the last 25 samples,
    \item the top 10 samples (according to the objective function),
    \item the top 20 samples,
    \item the top 30 samples.
\end{itemize}

\subsection{Training Models for Predicting ATPE Parameters}
\label{7-atpe-teaching}

ATPE introduces the concept of hyperparameter blocking, which optimizes the objective function within a reduced hyperparameter space. In this approach, some hyperparameters are fixed to constant values, simplifying the TPE search space during optimization.

Training models predicting ATPE parameters on real data, mentioned in the previous section, would require vast computational resources, as it is a time-consuming process. In an optimistic scenario where LightGBM models could learn parameter predictions in 1,000 iterations, ATPE optimization for a single problem would require about 100 iterations, and consequently, for a model with a one-minute training time per iteration, the entire training process would take over two months.

It’s important to note that Bayesian optimization methods are primarily applied to complex models because they yield better results with a limited number of objective function evaluations. Achieving effective ATPE parameter prediction models for these complex optimization tasks would require at least partial training on these models, potentially extending the computation time to several decades.

This issue can be addressed by substituting the objective function based on full model training with its surrogate. Such a surrogate function, like the original objective in HPO, takes hyperparameters as input but computes a real-valued function with at most a few hundred calculations. This reduces the computation time from several minutes or hours to just a few milliseconds, allowing for comprehensive training of the LightGBM models in a matter of hours.

The remaining challenge is to construct a class of functions that can be computed in a few hundred operations while closely resembling real objective functions in hyperparameter space. In ATPE these functions are designed as combinations of specific unary functions for each hyperparameter and binary functions for hyperparameter pairs~\cite{arsenault2023optimizing}. These binary functions simulate interdependencies between hyperparameters, which are often observed in real models.

\begin{figure}
\centering
\fbox{\includegraphics[width=0.4\textwidth]{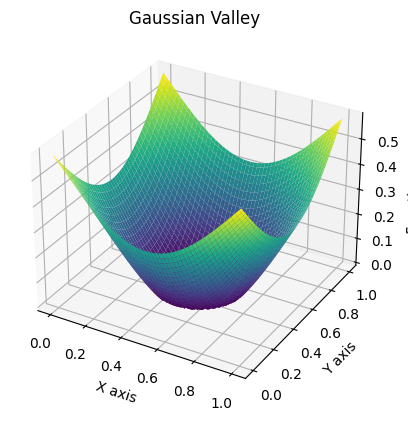}}
\caption{Simulated function - Gaussian product.}
\label{fig:sim-gauss}
\end{figure}

\begin{figure}
\centering
\fbox{\includegraphics[width=0.4\textwidth]{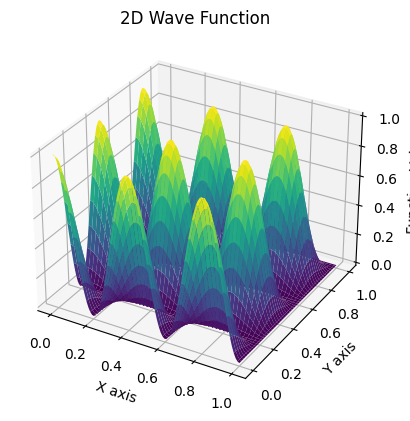}}
\caption{Simulated function - sine product.}
\label{fig:sim-sin}
\end{figure}

Another ATPE enhancement to the training process of parameter prediction models consists in calculating the statistics after generating thousands of simulated objective functions, and feeding them into the model (see Section \ref{7-atpe-stats}). Then a clustering algorithm groups these simulated functions based on similar statistical profiles, with only randomly selected representatives from each cluster used in further training. This technique ensures high data diversity during training while significantly reducing computational cost.

\section{Modifications to ATPE}

This chapter presents our proposed modifications to ATPE. Some of these changes are speculative, while others are grounded in an analysis of the model’s dynamics during benchmark function optimization. The modifications mainly aim to introduce new, potentially useful mechanisms to ATPE. As with those in the original work, their selection during the optimization process depends on the decisions made by ATPE’s LightGBM-based parameter prediction model. If this model frequently selects the new mechanisms, it can be inferred that they contribute significantly to the overall improvement of ATPE’s effectiveness, as the model generally prioritizes configurations that are likely to yield better results in a long term.

Following a discussion presented in Section~\ref{atpe-sect} we identified four key areas of potential ATPE improvements. These are: (1) different interpretation of \textit{secondaryCutoff} based on reversing the sequences for positive and negative values (discussed in Section~\ref{sec-cutoff}), (2) proposition of novel filtering schemes (Section~\ref{filt-sec}), (3) adding new components in the surrogate objective functions (Section~\ref{new-simulated}), and (4) introduction of a mechanism for blocking categorical hyperparameters (Section~\ref{mod-badging}).

The evaluation of the proposed ATPE modifications relies on benchmark functions commonly used to assess HPO algorithms. Direct optimization of ML models, especially with repeated training to ensure statistical reliability, would be computationally prohibitive. Instead, benchmark functions simulate objective function values within the hyperparameter space, allowing for controlled and repeatable testing of optimization algorithms. This approach is akin to the evaluation methodology in classical optimization algorithms, with particular emphasis on minimizing the number of objective function evaluations—a key factor in real-world HPO scenarios. We present comparisons against ATPE rather than TPE because, in our tests, ATPE outperformed TPE on every benchmark (see Table \ref{tab:tpe_atpe_comparison}).

\begin{table}[htbp]
\centering
\renewcommand{\arraystretch}{1.3}
\begin{tabular}{|l|c|c|c|c|}
\hline
 & \multicolumn{2}{c|}{\textbf{Average}} & \multicolumn{2}{c|}{\textbf{Median}} \\
\hline
\textbf{Benchmark} & \textbf{TPE} & \textbf{ATPE} & \textbf{TPE} & \textbf{ATPE} \\
\hline
Bohachevsky    & 116.696 & \cellcolor{red!20}24.044 & 59.199 & \cellcolor{red!20}7.857 \\
\hline
Branin         & 0.991   & \cellcolor{red!20}0.541  & 0.774  & \cellcolor{red!20}0.455 \\
\hline
Camelback      & -0.826  & \cellcolor{red!20}-1.000 & -0.895 & \cellcolor{red!20}-1.026 \\
\hline
Forrester      & -5.994  & \cellcolor{red!20}-6.019 & -6.018 & \cellcolor{red!20}-6.021 \\
\hline
GoldsteinPrice & 14.217  & \cellcolor{red!20}6.285  & 8.896  & \cellcolor{red!20}3.586 \\
\hline
Hartmann3      & -3.695  & \cellcolor{red!20}-3.820 & -3.756 & \cellcolor{red!20}-3.850 \\
\hline
Hartmann6      & -2.506  & \cellcolor{red!20}-2.969 & -2.598 & \cellcolor{red!20}-3.079 \\
\hline
Levy           & 0.023   & \cellcolor{red!20}0.003  & 0.001  & \cellcolor{red!20}0.000 \\
\hline
Rosenbrock     & 8.938   & \cellcolor{red!20}3.617  & 3.896  & \cellcolor{red!20}0.813 \\
\hline
\end{tabular}
\vspace{1em}
\caption{Mean and median objective values for benchmarks using ATPE and TPE (lower values are better).}
\label{tab:tpe_atpe_comparison}
\end{table}

Nine benchmark functions available in the HPOlib 1.5 library~\cite{HPOlib} were selected for analysis. These include both lower-dimensional functions like Forrester and Hartmann 3D~\cite{jones1998efficient}, and more complex, multimodal functions like Goldstein-Price and Levy~\cite{simulationlib}. Previous work with ATPE demonstrated its strong performance on some of these benchmarks, notably Hartmann 3D and Hartmann 6D, where it generally surpasses TPE, while achieving less consistent results on functions such as Rosenbrock and Goldstein-Price. The use of a diverse set of benchmarks provides insights into how each proposed ATPE modification performs across different types of optimization landscapes.

\subsection{Addressing the problem of "Discontinuity" of the Parameter Selecting Hyperparameters to Block}
\label{sec-cutoff}

The hyperparameter blocking process in ATPE operates in two stages. The first stage selects hyperparameters to be blocked based on Spearman’s correlation between hyperparameter values and corresponding objective function values. In the second stage, specific values are assigned to the blocked hyperparameters according to the current strategy.

In the original ATPE solution, the selection of hyperparameters is controlled by the parameters \textit{secondaryCutoff} and \textit{secondaryCorrelationExponent}. The latter one adjusts the Spearman correlation values, while the former one determines which and how many hyperparameters are candidates for blocking. The selection pseudo-algorithm is as follows:

\begin{enumerate}
    \item Let $n$ represent the total number of hyperparameters.
    \item If \textit{secondaryCutoff} $<0$, select $\lfloor|\textit{secondaryCutoff}|\cdot n \rfloor$ hyperparameters with the highest absolute correlation values for blocking.
    \item If \textit{secondaryCutoff} $>0$, select $\lfloor|\textit{secondaryCutoff}| \cdot n \rfloor$ hyperparameters with the lowest absolute correlation values for blocking.
\end{enumerate}

\begin{figure}[H]
\centering
\fbox{\includegraphics[width=0.9\textwidth]{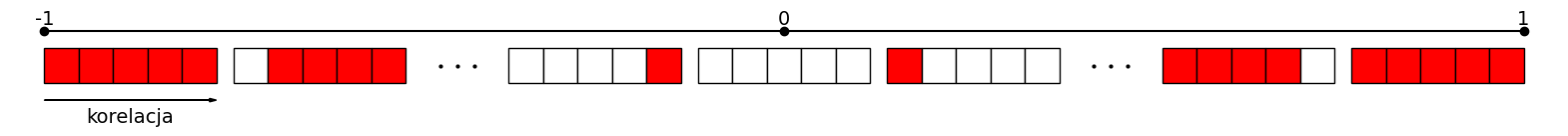}}
\caption{Visualization of hyperparameters selected for blocking based on \textit{secondaryCutoff} values in ATPE.}
\label{fig:org_cutoff}
\end{figure}

The \textit{secondaryCutoff} parameter ranges from $[-1,1]$. Hyperparameters selected based on this parameter follow a natural pattern, where each successive set includes or removes a hyperparameter according to the correlation order (see Figure \ref{fig:org_cutoff}).

Assuming objective function values change predictably with blocked hyperparameter adjustments, we can expect a similarly predictable response from the objective function with respect to \textit{secondaryCutoff}. This property often simplifies model building by avoiding the need to model more complex relationships.

\begin{figure}[H]
\centering
\fbox{\includegraphics[width=0.9\textwidth]{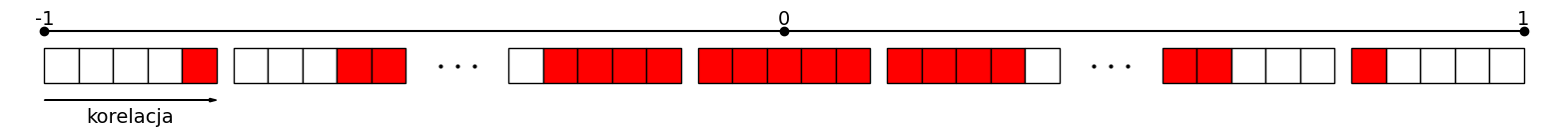}}
\caption{Visualization of hyperparameters selected for blocking based on \textit{secondaryCutoff} values in ATPE-r.}
\label{fig:mod_cutoff}
\end{figure}

However, a discontinuity in \textit{secondaryCutoff} arises around zero. With $n$ hyperparameters, a \textit{secondaryCutoff} of $\frac{1}{n}$ results in a set containing only the hyperparameter with the lowest correlation, while a \textit{secondaryCutoff} of $-\frac{1}{n}$ results in a set with only the highest correlation. 
This “discontinuity” can be resolved by slightly modifying the interpretation of \textit{secondaryCutoff} so that the sequences selected for positive and negative values are reversed (see Figure~\ref{fig:mod_cutoff}).

\begin{figure}[H]
\centering
\fbox{\includegraphics[width=0.7\textwidth]{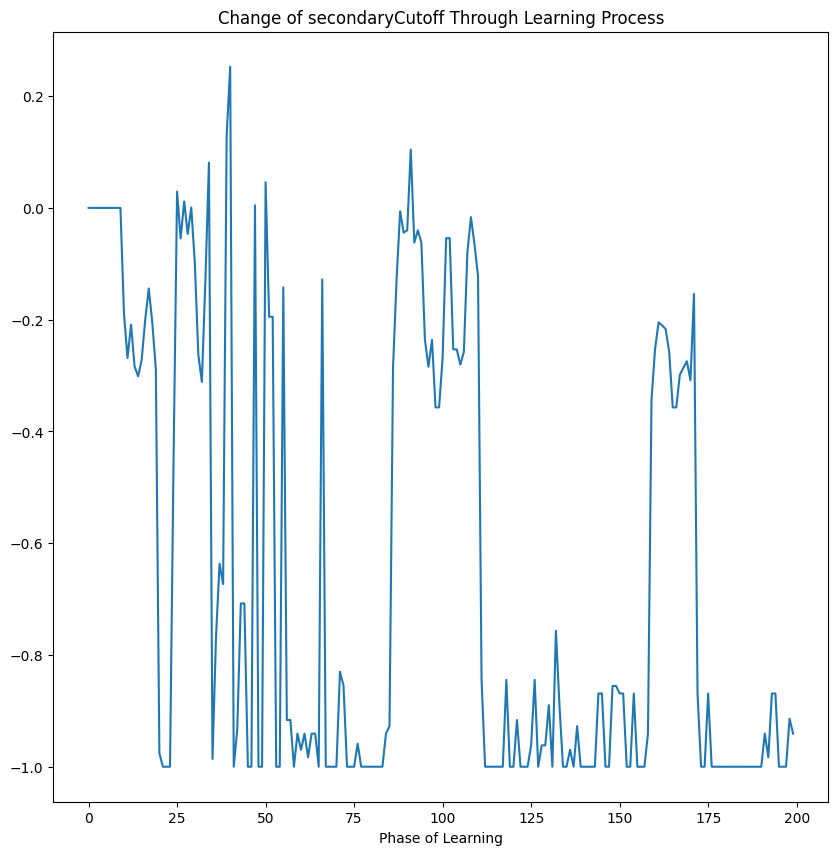}}
\caption{The progression of \textit{secondaryCutoff} values in one optimization using ATPE.}
\label{fig:proc_cutoff}
\end{figure}

This modification (ATPE-r) can lead to a smoother \textit{secondaryCutoff} progression during training (see Figure \ref{fig:proc_cutoff}), enhancing the effectiveness of hyperparameter blocking throughout optimization. The pseudo-algorithm of ATPE-r \textit{secondaryCutoff} interpretation is as follows:
\begin{enumerate}
    \item Let $n$ represent the total number of hyperparameters.
    \item If \textit{secondaryCutoff} $<0$, select $n - \lfloor|\textit{secondaryCutoff}| \cdot n \rfloor$ hyperparameters with the highest absolute correlation values for blocking.
    \item If \textit{secondaryCutoff} $>0$, select $n - \lfloor|\textit{secondaryCutoff}| \cdot n \rfloor$ hyperparameters with the lowest absolute correlation values for blocking.
\end{enumerate}

\begin{figure}[H]
\centering
\fbox{\includegraphics[width=0.7\textwidth]{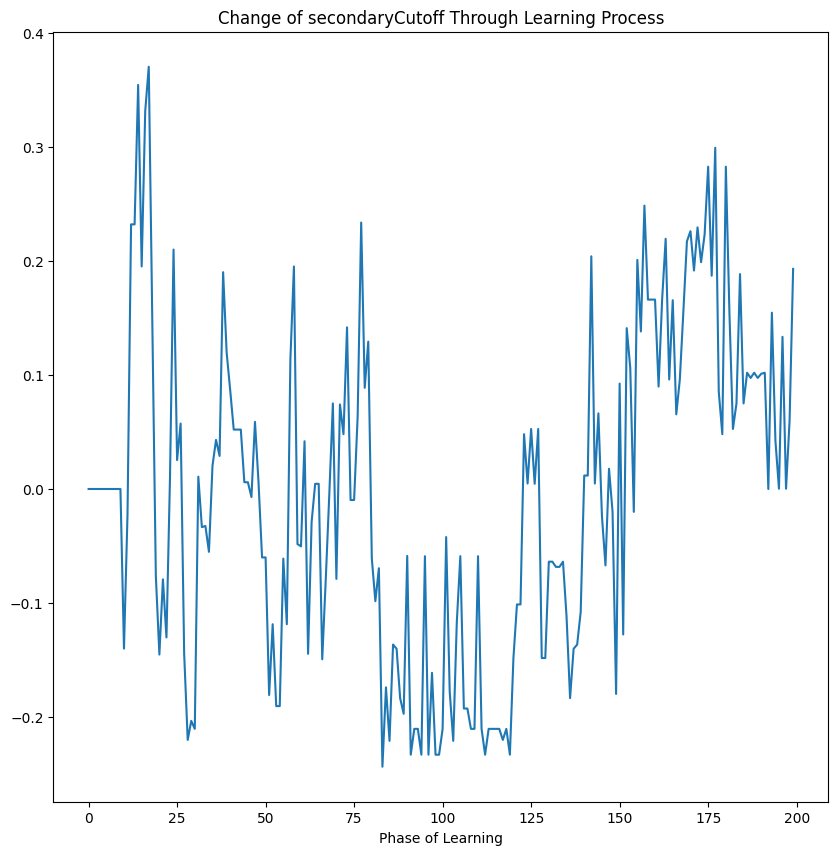}}
\caption{The progression of \textit{secondaryCutoff} values in one optimization using ATPE-r.}
\label{fig:new_cutoff}
\end{figure}

Table \ref{tab:continuity} compares the performance of the original ATPE solution and ATPE-r. In 4 out of 9 benchmarks, the ATPE-r model achieved a higher average, and in 7 out of 9, it achieved a higher median. This indicates that ATPE-r is more stable while maintaining a similar ability to find optimal solutions. The stabilization supports the hypothesis that the \textit{secondaryCutoff} discontinuity introduces some chaotic behavior. This conclusion is further confirmed by Figure \ref{fig:new_cutoff}, where the parameter values in ATPE-r vary more smoothly.

\begin{table}[htbp]
\centering
\renewcommand{\arraystretch}{1.3}
\begin{tabular}{|l|c|c|c|c|}
\hline
 & \multicolumn{2}{c|}{\textbf{Average}} & \multicolumn{2}{c|}{\textbf{Median}} \\
\hline
\textbf{Benchmark} & \textbf{ATPE} & \textbf{ATPE-r} & \textbf{ATPE} & \textbf{ATPE-r} \\
\hline
Bohachevsky   & \cellcolor{red!20}24.044 & 24.356 & 7.857 & \cellcolor{red!20}6.626 \\
\hline
Branin        & 0.541 & \cellcolor{red!20}0.536 & 0.455 & \cellcolor{red!20}0.417 \\
\hline
Camelback     & -1.000 & \cellcolor{red!20}-1.003 & -1.026 & -1.026 \\
\hline
Forrester     & \cellcolor{red!20}-6.019 & -6.016 & -6.021 & -6.021 \\
\hline
GoldsteinPrice& 6.285 & \cellcolor{red!20}6.242 & 3.586 & \cellcolor{red!20}3.289 \\
\hline
Hartmann3     & \cellcolor{red!20}-3.820 & -3.809 & -3.850 & \cellcolor{red!20}-3.853 \\
\hline
Hartmann6     & \cellcolor{red!20}-2.969 & -2.900 & \cellcolor{red!20}-3.079 & -3.068 \\
\hline
Levy          & \cellcolor{red!20}0.003 & 0.005 & 0.000 & 0.000 \\
\hline
Rosenbrock    & 3.617 & \cellcolor{red!20}2.916 & \cellcolor{red!20}0.813 & 1.105 \\
\hline
\end{tabular}
\vspace{1em}
\caption{Mean and median benchmark objective values for the original ATPE and modified \textit{secondaryCutoff} (the lower the values the better).}
\label{tab:continuity}
\end{table}

\subsection{Novel Filtering Schemes}
\label{filt-sec}

\begin{figure}[H]
\centering
\fbox{\includegraphics[width=0.8\textwidth]{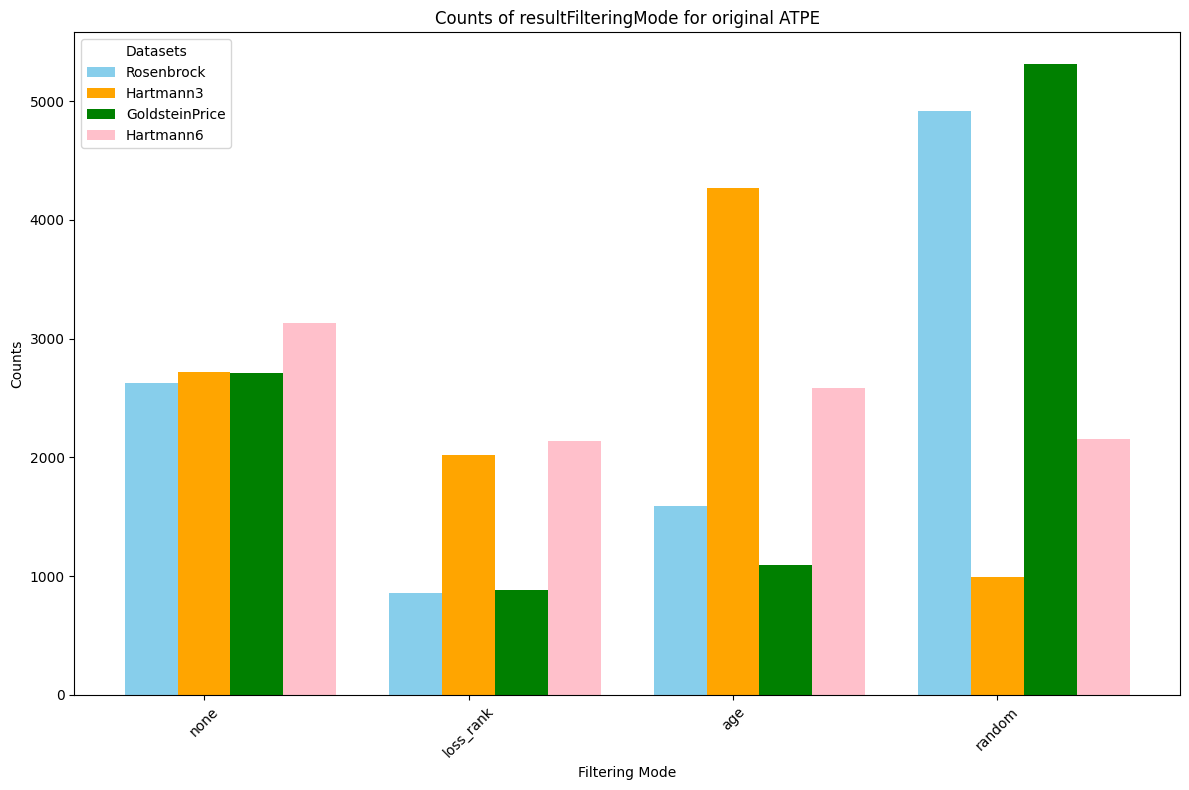}}
\caption{Frequency distribution of filtering schemes used by ATPE on selected benchmarks.}
\label{fig:filtering_old}
\end{figure}

ATPE filters out part of the sample history before passing it to TPE (see Section \ref{7-atpe-filt}). This section examines the dynamics of existing filtering schemes with the potential to introduce new ones. The benchmarks selected for this analysis include Hartmann 3D, Hartmann 6D, Rosenbrock, and Goldstein-Price functions. ATPE outperforms TPE on the first two benchmarks and generally performs similarly on the remaining ones. In contrast, Rosenbrock and Goldstein-Price are the only benchmarks where ATPE fails to outperform TPE. Interestingly, in this study, ATPE does not significantly outperform TPE on Hartmann 3D and Hartmann 6D functions. Figure \ref{fig:filtering_old} shows the frequency distribution of filtering schemes chosen by LightGBM over 50 rounds of four 200-step benchmarks. Notably, the choice of filtering parameters in the model differs significantly between the pairs (Hartmann 3D, Hartmann 6D) and (Rosenbrock, Goldstein-Price). In the latter one, LightGBM selects the objective-based scheme less frequently and relies more on random and age-based filtering. This adjustment decreases exploitation, shifting the balance more toward exploration. These functions have few local minima, making this behavior unexpected, as optimization algorithms typically favor exploration when optimizing highly multimodal functions to locate better areas for further exploitation.

This observation led us to the hypothesis that ATPE may find the objective-based filtering scheme ineffective and therefore avoids it. Considering potential modifications to the filtering scheme to improve model performance, the high frequency of random selection also suggests a potential opportunity for improvement. Introducing a more exploration-focused scheme could prompt the model to replace the random option with this new scheme.

Two additional filtering schemes were proposed in this study:

(1): \textit{Clustering-based}: The sample history is divided into clusters based on the numerical values of hyperparameters using the K-means method. The number of clusters is determined by a new parameter, \emph{clustersQuantile}, multiplied by the length of the history. One representative is then selected from each cluster.

(2): \textit{Z-score-based}: A Z-score standardization value is calculated for all samples in the history, given by 
\[
z = \frac{x- \mu}{\sigma}
\]
where $\mu$ is the mean objective function value, and $\sigma$ is the standard deviation. Samples are selected based on a threshold value, \emph{zscoreThreshold}, according to the following scheme:
\begin{enumerate}
    \item if $\emph{zscoreThreshold} < 0$, select samples belonging to:
    \[
    \{x: zscore(x) > |\emph{zscoreThreshold}|\}
    \]
    \item otherwise, select samples belonging to:
    \begin{equation}
    \{x: zscore(x) < 3 - |\emph{zscoreThreshold}|\}
    \label{eq:zscore_t}
    \end{equation}
\end{enumerate}

The first of these options introduces a more intensive exploration scheme than random filtering, as the representatives from each cluster form a diverse subset of samples. The second option selects the most or least extreme samples based on the parameter's sign, serving both exploratory and exploitative strategies. The threshold for positive parameter values is set to $3 - |\emph{zscoreThreshold}|$, instead of simply $|\emph{zscoreThreshold}|$, to maintain continuity, similarly to the modification described in Section \ref{sec-cutoff}. Although Z-scores can theoretically exceed 3, they rarely do in practice, hence the use of this value in the threshold expression~\ref{eq:zscore_t}.

\begin{figure}[H]
\centering
\fbox{\includegraphics[width=0.8\textwidth]{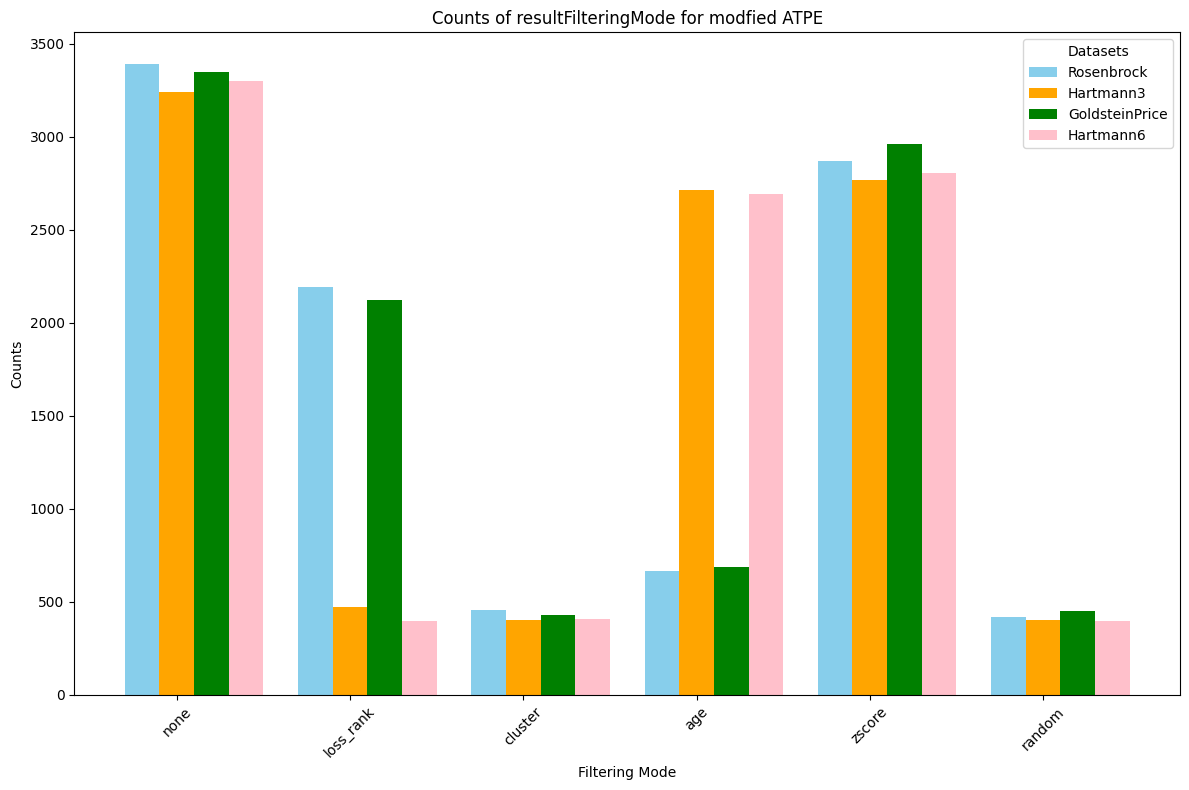}}
\caption{Frequency distribution of filtering schemes used by ATPE-f on selected benchmarks.}
\label{fig:filtering_new}
\end{figure}

Figure \ref{fig:filtering_new} shows the frequency distribution of the modified filtering options over a 50-round, 200-step benchmark. The modifications significantly altered the model's behavior, with random filtering used much less frequently. Of the new options, the Z-score-based one proved more effective, as indicated by its frequent selection by the model. The clustering-based option was rarely chosen, suggesting lower effectiveness.

Table \ref{tab:filtering_comparison} compares the original ATPE and modified algorithm (ATPE-f) performance across all benchmarks. Although the results are similar on most benchmarks, ATPE gives better average results on 6 out of 9 benchmarks. Improved results of ATPE-f are observed for Hartmann 6, Levy, and Rosenbrock functions. The Rosenbrock function is particularly notable, as it was one of the few benchmarks where the original TPE outperformed ATPE. ATPE-f achieved significantly better results on this benchmark than ATPE.

\begin{table}[htbp]
\centering
\renewcommand{\arraystretch}{1.3}
\begin{tabular}{|l|c|c|c|c|}
\hline
 & \multicolumn{2}{c|}{\textbf{Average}} & \multicolumn{2}{c|}{\textbf{Median}} \\
\hline
\textbf{Benchmark} & \textbf{ATPE} & \textbf{ATPE-f} & \textbf{ATPE} & \textbf{ATPE-f} \\
\hline
Bohachevsky 
  & \cellcolor{red!20}24.044 & 26.269 
  & 7.857 & \cellcolor{red!20}4.488 \\
\hline
Branin 
  & \cellcolor{red!20}0.541 & 0.552
  & 0.455 & \cellcolor{red!20}0.440 \\
\hline
Camelback 
  & \cellcolor{red!20}-1.000 & -0.992
  & \cellcolor{red!20}-1.026 & -1.017 \\
\hline
Forrester 
  & \cellcolor{red!20}-6.019 & -6.017
  & \cellcolor{red!20}-6.021 & -6.020 \\
\hline
GoldsteinPrice 
  & \cellcolor{red!20}6.285 & 6.915
  & 3.586 & \cellcolor{red!20}3.403 \\
\hline
Hartmann3 
  & \cellcolor{red!20}-3.820 & -3.816
  & \cellcolor{red!20}-3.850 & -3.846 \\
\hline
Hartmann6 
  & -2.969 & \cellcolor{red!20}-2.998
  & -3.079 & \cellcolor{red!20}-3.130 \\
\hline
Levy 
  & 0.003 & \cellcolor{red!20}0.002
  & 0.000 & 0.000 \\
\hline
Rosenbrock 
  & 3.617 & \cellcolor{red!20}3.035
  & 0.813 & \cellcolor{red!20}0.679 \\
\hline
\end{tabular}
\vspace{1em}
\caption{Mean and median objective values for benchmarks using ATPE and ATPE-f filtering schemes (lower values are better).}
\label{tab:filtering_comparison}
\end{table}

\subsection{New Components in Surrogate Objective Functions}
\label{new-simulated}

Parameter-predicting models in ATPE are trained using surrogate objective functions, constructed as sums of certain unary and binary functions (cf. Section~\ref{7-atpe-teaching}). Aiming to investigate whether expanding the set of possible component functions would result in more effective parameter-predicting models, and consequently improve the ATPE benchmark performance, we introduced two new component functions:

(1) \textit{Sigmoid} – a unary, classic sigmoid function, defined as 
\[
f(h_i) = \frac{1}{1+e^{-a(h_i-b)}},
\]
where $a$ controls the slope, and $b$ the point at which the function starts to rise steeply. This function is intended to introduce smooth, nonlinear curves, and is based on the observation that in ML models, certain parameters may significantly affect the objective function value up to a certain threshold, beyond which their influence becomes negligible (e.g., learning rate \cite{bello2017neuraloptimizersearchreinforcement}).

\begin{figure}[H]
\centering
\fbox{\includegraphics[width=0.5\textwidth]{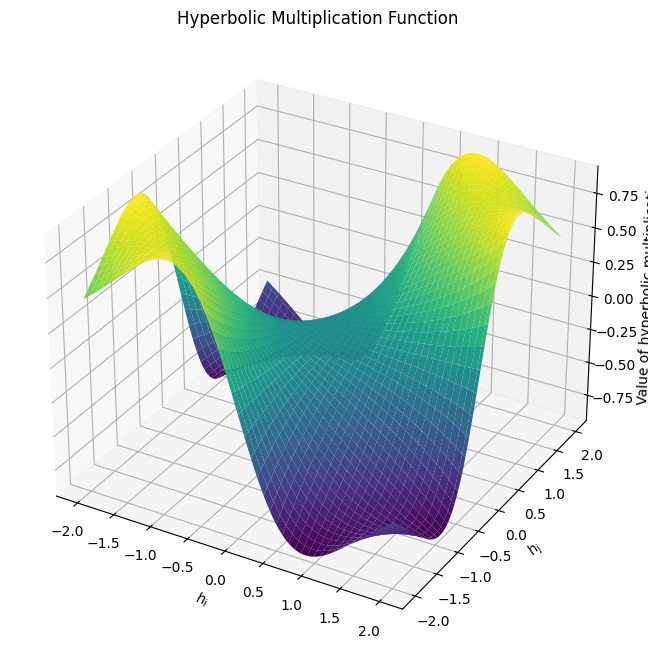}}
\caption{Hyperbolic product}
\label{fig:hyperbolic}
\end{figure}

(2) \textit{Hyperbolic product} – a binary function, defined as
\[
f(h_i,h_j)  = \frac{\sinh(ah_i) \cdot \sinh(bh_j)}{c+\cosh(h_ih_j)},
\]
where parameters $a$, $b$, and $c$ are randomly assigned. This function introduces a new type of nonlinearity with sharper gradients (see Figure~\ref{fig:hyperbolic}).

Table \ref{tab:new_sim_comparison} compares the original ATPE with the modified ATPE incorporating the two above component functions in the surrogate objective function (ATPE-cf). ATPE-cf outperformed ATPE on most benchmarks, with only 3 out of 9 showing inconclusive results (one model with a better mean and the other one with a better median). The largest improvement was observed in the Rosenbrock benchmark, where ATPE-cf achieved better results than models with the \textit{secondaryCutoff} and filtering scheme modifications.

\begin{table}[htbp]
\centering
\renewcommand{\arraystretch}{1.3}
\begin{tabular}{|l|c|c|c|c|}
\hline
 & \multicolumn{2}{c|}{\textbf{Average}} & \multicolumn{2}{c|}{\textbf{Median}} \\
\hline
\textbf{Benchmark} & \textbf{ATPE} & \textbf{ATPE-cf} & \textbf{ATPE} & \textbf{ATPE-cf}\\
\hline
Bohachevsky    & \cellcolor{red!20}24.044 & 24.115 & 7.857 & \cellcolor{red!20}2.402 \\
\hline
Branin         & 0.541 & \cellcolor{red!20}0.489 & 0.455 & \cellcolor{red!20}0.412 \\
\hline
Camelback      & -1.000 & \cellcolor{red!20}-1.000  & \cellcolor{red!20}-1.026 & -1.025 \\
\hline
Forrester      & \cellcolor{red!20}-6.019 & -6.016 & -6.021 & \cellcolor{red!20}-6.021  \\
\hline
GoldsteinPrice & \cellcolor{red!20}6.285 & 6.796 & 3.586 & \cellcolor{red!20}3.196 \\
\hline
Hartmann3      & -3.820 & \cellcolor{red!20}-3.822 & -3.850 & \cellcolor{red!20}-3.853 \\
\hline
Hartmann6      & -2.969 & \cellcolor{red!20}-2.995 & -3.079 & \cellcolor{red!20}-3.151 \\
\hline
Levy         & 0.003  & \cellcolor{red!20}0.003 & 0.000 & \cellcolor{red!20}0.000  \\
\hline
Rosenbrock     & 3.617 & \cellcolor{red!20}1.683 & 0.813 & \cellcolor{red!20}0.589 \\
\hline
\end{tabular}
\vspace{1em}
\caption{Mean and median objective values for benchmarks using ATPE and ATPE-cf (lower values are better).}
\label{tab:new_sim_comparison}
\end{table}

\subsection{Alternative Models for Predicting ATPE Hyperparameters}
\label{different-models}

In the original ATPE study, LightGBM was chosen for hyperparameter prediction (see Sections \ref{7-atpe-optim}) due to its demonstrated effectiveness in complex tasks. However, other architectures may also be suited to this role. This study explores the performance of two alternatives: CatBoost and SVM.

CatBoost, introduced in 2019 \cite{prokhorenkova2019catboostunbiasedboostingcategorical}, offers strong handling of categorical variables, several of which are present in ATPE. It also adapts well to small datasets and complex variable dependencies, making it a promising candidate.

For SVM, the RBF kernel was used. Since SVM does not inherently calculate feature importance, SHAP was employed to estimate feature impacts required by certain parts of the ATPE algorithm \cite{lundberg2017unifiedapproachinterpretingmodel}. Due to the computational expense of SHAP, benchmark rounds were reduced to 5 (from 50), which may increase error margins in the results.

Table \ref{tab:catboost_comp} compares the original ATPE with ATPE using CatBoost (ATPE-cb). CatBoost produced comparable or slightly worse results across most benchmarks. Medians were marginally better in 5 out of 9 benchmarks, suggesting similar stability.

Table \ref{tab:svm_comp} compares the original ATPE with ATPE using SVM (ATPE-sv). The modified model generally performed worse across benchmarks, with only a few results being close to the original. SVM did not outperform LightGBM in any case, though reduced benchmark rounds may affect the comparability.

\begin{table}[htbp]
\centering
\renewcommand{\arraystretch}{1.3}
\begin{tabular}{|l|c|c|c|c|}
\hline
 & \multicolumn{2}{c|}{\textbf{Average}} & \multicolumn{2}{c|}{\textbf{Median}} \\
\hline
\textbf{Benchmark} & \textbf{ATPE} & \textbf{ATPE-cb} & \textbf{ATPE} & \textbf{ATPE-cb}\\
\hline
Bohachevsky    & \cellcolor{red!20}24.044 & 35.509 & \cellcolor{red!20}7.857 & 11.012 \\
\hline
Branin         & \cellcolor{red!20}0.541 & 0.585 & \cellcolor{red!20}0.455 & 0.471 \\
\hline
Camelback      & \cellcolor{red!20}-1.000 & -0.998 & -1.026 &\cellcolor{red!20}-1.026  \\
\hline
Forrester      & \cellcolor{red!20}-6.019 & -6.014 & -6.021 & \cellcolor{red!20}-6.021  \\
\hline
GoldsteinPrice & \cellcolor{red!20}6.285 & 6.422 & \cellcolor{red!20}3.586 & 3.999 \\
\hline
Hartmann3      & \cellcolor{red!20}-3.820 & -3.816 & -3.850 & \cellcolor{red!20}-3.855 \\
\hline
Hartmann6      & \cellcolor{red!20}-2.969 & -2.942 & -3.079 & \cellcolor{red!20}-3.094 \\
\hline
Levy           & \cellcolor{red!20}0.003 & 0.004 & 0.000 & \cellcolor{red!20}0.000  \\
\hline
Rosenbrock     & 3.617 & \cellcolor{red!20}2.825 & \cellcolor{red!20}0.813 & 1.181 \\
\hline
\end{tabular}
\vspace{1em}
\caption{Mean and median objective values for benchmarks using the original ATPE and ATPE with CatBoost as the parameter prediction model (lower values are better).}
\label{tab:catboost_comp}
\end{table}

\begin{table}[htbp]
\centering
\renewcommand{\arraystretch}{1.3}
\begin{tabular}{|l|c|c|c|c|}
\hline
 & \multicolumn{2}{c|}{\textbf{Average}} & \multicolumn{2}{c|}{\textbf{Median}} \\
\hline
\textbf{Benchmark} & \textbf{ATPE} & \textbf{ATPE-sv} & \textbf{ATPE} & \textbf{ATPE-sv} \\
\hline
Bohachevsky    & \cellcolor{red!20}24.044 & 142.991 & \cellcolor{red!20}7.857 & 23.484 \\
\hline
Branin         & \cellcolor{red!20}0.541 & 1.197   & \cellcolor{red!20}0.455 & 0.602 \\
\hline
Camelback      & \cellcolor{red!20}-1.000 & -0.801  & \cellcolor{red!20}-1.026 & -0.968 \\
\hline
Forrester      & \cellcolor{red!20}-6.019 & -5.764  & \cellcolor{red!20}-6.021 & -6.012 \\
\hline
GoldsteinPrice & \cellcolor{red!20}6.285 & 62.130  & \cellcolor{red!20}3.586 & 5.846 \\
\hline
Hartmann3      & \cellcolor{red!20}-3.820 & -3.640  & \cellcolor{red!20}-3.850 & -3.792 \\
\hline
Hartmann6      & \cellcolor{red!20}-2.969 & -2.609  & \cellcolor{red!20}-3.079 & -2.887 \\
\hline
Levy           & \cellcolor{red!20}0.003 & 0.092   & \cellcolor{red!20}0.000 & 0.009 \\
\hline
Rosenbrock     & \cellcolor{red!20}3.617 & 294.470 & \cellcolor{red!20}0.813 & 4.402 \\
\hline
\end{tabular}
\vspace{1em}
\caption{Mean and median objective values for benchmarks using the original ATPE and ATPE with SVM as the parameter prediction model (lower values are better).}
\label{tab:svm_comp}
\end{table}

\subsection{Modification of the Hyperparameter Blocking Scheme}
\label{mod-badging}

\sloppy
The hyperparameter blocking process in ATPE is limited to non-categorical variables due to its reliance on Spearman's correlation 
(eq.~\ref{eq:Spearman}). In this section we propose a mechanism for blocking categorical hyperparameters, as well. The general logic remains consistent with the original ATPE design, except that categorical parameter decisions are now based on ANOVA results rather than Spearman's correlation. New parameters are added to regulate this blocking strategy, and the continuity modification described in \ref{sec-cutoff} is applied. A pseudo-algorithm of theis process is outlined below.

For each categorical hyperparameter, the ANOVA value with respect to the objective function is calculated. A weighted ANOVA value is then computed as
\begin{equation}
ANOVA'(x) = |ANOVA(x)|^\alpha
\end{equation}
where $\alpha$ is a model parameter.
The hyperparameters are sorted in descending order of $ANOVA'(x)$ to create the sequence $a_{i=1,..,K}$.
Next, the weighted ANOVA sum for all hyperparameters is calculated and multiplied by the absolute value of the threshold $\beta$, which is a model parameter:
\begin{equation}
T = \left(\sum_{i=1}^K ANOVA'(a_i)\right) \cdot (1 - |\beta|) 
\end{equation}
If $\beta > 0$, we select hyperparameters with the highest weighted ANOVA values until their sum meets the threshold:
\begin{equation}
W = \{i: G(i) \leq  T, \quad 1 \leq i \leq K\}
\end{equation}
where
\begin{equation}
G(i) = ANOVA'(a_{i}) + G(i-1), \quad G(0) = 0 \land 0 \leq i \leq K
\end{equation}
If $\beta < 0$, we select hyperparameters with the lowest weighted ANOVA values until their sum meets the threshold:
\begin{equation}
W = \{i: L(i) \leq T,  \quad 1 \leq i \leq K\}
\end{equation}
where
\begin{equation}
L(i) = ANOVA'(a_{i}) + L(i+1), \quad L(K+1) = 0 \land 1 \leq i \leq K + 1
\end{equation}
Parameters $\alpha$ and $\beta$, referred to as \textit{secondaryAnovaExponent} and \textit{secondaryCatCutoff}, are new model parameters predicted by LightGBM.

The remaining selection process proceeds as for numerical variables, i.e., from the hyperparameters selected in the first stage, parameters are randomly chosen based on either:
\begin{itemize}
    \item a fixed probability (similarly to continuous variables),
    \item a probability based on the weighted ANOVA value (as calculated above) multiplied by a new model parameter (\textit{secondaryAnovaMultiplier}). 
\end{itemize}

Table \ref{tab:cat_comp} shows a comparison between the original ATPE and the modified ATPE with categorical variables included in the blocking scheme (ATPE-c). ATPE-c achieved better medians in 7 out of 9 benchmarks, though on average it outperformed the original model in only 3 out of 9 benchmarks. 

Overall, the differnces were generally small, except foir the Bohachevsky function (in both average and median values), GoldsteinPrice (in terms of averages), and Rosenbrock function (average values).

\begin{table}[htbp]
\centering
\renewcommand{\arraystretch}{1.3}
\begin{tabular}{|l|c|c|c|c|}
\hline
 & \multicolumn{2}{c|}{\textbf{Average}} & \multicolumn{2}{c|}{\textbf{Median}} \\
\hline
\textbf{Benchmark} & \textbf{ATPE} & \textbf{ATPE-c} & \textbf{ATPE} & \textbf{ATPE-c}\\
\hline
Bohachevsky    & \cellcolor{red!20}24.044 & 35.285 & 7.857 & \cellcolor{red!20}6.402 \\
\hline
Branin         & 0.541 & \cellcolor{red!20}0.536 & 0.455 & \cellcolor{red!20}0.438 \\
\hline
Camelback      & \cellcolor{red!20}-1.000 & -0.996 & \cellcolor{red!20}-1.026 & -1.023 \\
\hline
Forrester      & \cellcolor{red!20}-6.019 & -6.013 & -6.021 & \cellcolor{red!20}-6.021 \\
\hline
GoldsteinPrice & \cellcolor{red!20}6.285 & 8.205  & 3.586 & \cellcolor{red!20}3.538 \\
\hline
Hartmann3      & \cellcolor{red!20}-3.820 & -3.815 & -3.850 & \cellcolor{red!20}-3.850 \\
\hline
Hartmann6      & -2.969 & \cellcolor{red!20}-2.991 & -3.079 & \cellcolor{red!20}-3.136 \\
\hline
Levy           & \cellcolor{red!20}0.003 & 0.007  & 0.000 & \cellcolor{red!20}0.000 \\
\hline
Rosenbrock     & 3.617 & \cellcolor{red!20}2.125  & \cellcolor{red!20}0.813 & 0.868 \\
\hline
\end{tabular}
\vspace{1em}
\caption{Mean and median objective values for benchmarks using ATPE and ATPE-c (lower values are better).}
\label{tab:cat_comp}
\end{table}

\section{Extended Results}
\subsection{Combination of the best-performing enhancements}

Before drawing general conclusions regarding the modifications introduced, additional training and benchmarking was conducted on a version of the algorithm incorporating all modifications proposed in the previous section (i.e. -r, -f, -cf, and -c) that were found to have a positive impact on the model performance. A modification -x (x $\in\{$r, f, cf, c$\}$) was considered positive if, compared to ATPE, the mean of ATPE-x was better in at least 5 out of 9 benchmarks and the median was better in at least 4 out of 9, or if the mean was better in at least 4 out of 9 and the median was better in at least 5 out of 9. There were two modifications meeting these criteria:

(1) -c: Discontinuity in the parameter for selecting hyperparameters to block (see Section~\ref{sec-cutoff}),

(2) -cf: New components in the surrogate objective functions (see Section~\ref{new-simulated}).

Table \ref{tab:cutoff_sim} compares the original ATPE with the modified ATPE-c-cf. It can be observed that these modifications seem to negatively interact with each other. In only 3 out of 9 benchmarks did the modified algorithm achieve a better mean, and it achieved a better median in 4 out of 9 benchmarks. It is difficult to speculate on the cause of this performance drop. Both modifications affect only the parameter-predicting model and not the ATPE algorithm’s primary structure.

\begin{table}[htbp]
\centering
\renewcommand{\arraystretch}{1.3}
\begin{tabular}{|l|c|c|c|c|}
\hline
 & \multicolumn{2}{c|}{\textbf{Average}} & \multicolumn{2}{c|}{\textbf{Median}} \\
\hline
\textbf{Benchmark} & \textbf{ATPE} & \textbf{ATPE-c-cf} & \textbf{ATPE} & \textbf{ATPE-c-cf}\\
\hline
Bohachevsky    & \cellcolor{red!20}24.044 & 40.924 & \cellcolor{red!20}7.857 & 10.346 \\
\hline
Branin         & 0.541 & \cellcolor{red!20}0.535 & 0.455 & \cellcolor{red!20}0.430 \\
\hline
Camelback      & \cellcolor{red!20}-1.000 & -0.995 & \cellcolor{red!20}-1.026 & -1.019 \\
\hline
Forrester      & \cellcolor{red!20}-6.019 & -6.016 & -6.021 & \cellcolor{red!20}-6.021 \\
\hline
GoldsteinPrice & \cellcolor{red!20}6.285 & 6.588  & 3.586 & \cellcolor{red!20}3.272 \\
\hline
Hartmann3      & \cellcolor{red!20}-3.820 & -3.796 & \cellcolor{red!20}-3.850 & -3.833 \\
\hline
Hartmann6      & \cellcolor{red!20}-2.969 & -2.877 & \cellcolor{red!20}-3.079 & -3.044 \\
\hline
Levy           & 0.003 & \cellcolor{red!20}0.003  & 0.000 & \cellcolor{red!20}0.000 \\
\hline
Rosenbrock     & 3.617 & \cellcolor{red!20}2.817 & \cellcolor{red!20}0.813 & 0.819 \\
\hline
\end{tabular}
\vspace{1em}
\caption{Mean and median objective values for benchmarks using the original ATPE and ATPE-c-cf, i.e., the modified version with \textit{secondaryCutoff} and new simulated function components (lower values are better).}
\label{tab:cutoff_sim}
\end{table}

\subsection{Evaluation of \textit{zscore} filtering}
Section~\ref{filt-sec} indicated potentially positive effects of the \textit{zscore} filtering method, which may have been overshadowed by the \textit{clustersQuantile} modification. This observation led us to test a model that used \textit{zscore} filtering along with the most effective modification—the new components for simulated objective functions (ATPE-cf-zscore). Table~\ref{tab:sim_zscore} compares these results with the original ATPE. ATPE-cf-zscore achieved better medians for all benchmarks and better means for 7 out of 9 benchmarks. Additionally, in one of the remaining two benchmark functions (Forrester), the difference between the models was minimal, suggesting that the observed advantage may be within statistical error.

Table~\ref{tab:best_for_each} presents the best-performing model for each benchmark. Notably, in 6 out of 9 benchmarks for mean and 8 out of 9 for median the top approach was either ATPE-cf or ATPE-cf-zscore, thus proving the relevance of adding the proposed new components to the surrogate evalution function. Although no single model outperformed the original ATPE on all benchmarks, it was possible to achieve better mean and median values for almost all benchmarks (the only exception being the Forrester function) using certain modification.

\begin{table}[htbp]
\centering
\renewcommand{\arraystretch}{1.3}
\begin{tabular}{|l|c|c|c|c|}
\hline
 & \multicolumn{2}{c|}{\textbf{Average}} & \multicolumn{2}{c|}{\textbf{Median}} \\
\hline
\textbf{Benchmark} & \textbf{ATPE} & \textbf{ATPE-cf-zscore} & \textbf{ATPE} & \textbf{ATPE-cf-zscore}\\
\hline
Bohachevsky
  & 24.044 & \cellcolor{red!20}20.597
  & 7.857  & \cellcolor{red!20}2.735 \\
\hline
Branin
  & 0.541 & \cellcolor{red!20}0.517
  & 0.455 & \cellcolor{red!20}0.410 \\
\hline
Camelback
  & -1.000 & \cellcolor{red!20}-1.003
  & -1.026 & \cellcolor{red!20}-1.026 \\
\hline
Forrester
  & \cellcolor{red!20}-6.019 & -6.016
  & -6.021 & \cellcolor{red!20}-6.021 \\
\hline
GoldsteinPrice
  & 6.285 & \cellcolor{red!20}5.720
  & 3.586 & \cellcolor{red!20}3.152 \\
\hline
Hartmann3
  & -3.820 & \cellcolor{red!20}-3.824
  & -3.850 & \cellcolor{red!20}-3.855 \\
\hline
Hartmann6
  & \cellcolor{red!20}-2.969 & -2.933
  & -3.079 & \cellcolor{red!20}-3.120 \\
\hline
Levy
  & 0.003 & \cellcolor{red!20}0.003
  & 0.000 & \cellcolor{red!20}0.000 \\
\hline
Rosenbrock
  & 3.617 & \cellcolor{red!20}2.479
  & 0.813 & \cellcolor{red!20}0.586 \\
\hline
\end{tabular}
\vspace{1em}
\caption{Mean and median objective values for benchmarks using ATPE and ATPE-cf-zscore (lower values are better).}
\label{tab:sim_zscore}
\end{table}

\begin{table}[htbp]
\renewcommand{\arraystretch}{1.3}
\centering
\begin{tabular}{|c|c|c|}
\hline
\textbf{Benchmark} & \textbf{Best Model (Average)} & \textbf{Best Model (Median)} \\
\hline
Bohachevsky    & APTE-cf-zscore & ATPE-cf \\
\hline
Branin         & ATPE-cf & APTE-cf-zscore \\
\hline
Camelback      & APTE-cf-zscore & ATPE-cb \\
\hline
Forrester      & ATPE & APTE-cf-zscore \\
\hline
GoldsteinPrice & APTE-cf-zscore & APTE-cf-zscore \\
\hline
Hartmann3      & APTE-cf-zscore & APTE-cf-zscore \\
\hline
Hartmann6      & ATPE-f  & ATPE-cf \\
\hline
Levy           & ATPE-f  & ATPE-cf \\
\hline
Rosenbrock     & ATPE-cf & APTE-cf-zscore \\
\hline
\end{tabular}
\vspace{1em}
\caption{Best-performing modification for each benchmark in terms of mean and median.}
\label{tab:best_for_each}
\end{table}

\section{Conclusions}
This study introduced various modifications to the ATPE algorithm to improve its performance. The changes implemented had diverse impacts on the effectiveness of the algorithm, indicating that there is no universal solution for all benchmarks. Some modifications proved more effective, suggesting potential for further optimization by strategically combining different approaches. The experimental results highlight the importance of trying various strategies and analyzing their impacts on specific optimization tasks.

\section*{Acknowledgments}
This work was supported by the National Science Centre, grant number 2023/49/B/ST6/01404.

\bibliographystyle{unsrt}
\bibliography{references}

\end{document}